\documentclass[letterpaper, 10 pt, conference]{IEEEtran} % Comment this line out if you need a4paper

\IEEEoverridecommandlockouts     % This command is only needed if 
\usepackage{cite}
\usepackage{graphics} % for pdf, bitmapped graphics files
\usepackage{epsfig} % for postscript graphics files
\usepackage{times} % assumes new font selection scheme installed

\usepackage{xcolor}
\usepackage{amsmath} % assumes amsmath package installed
\usepackage{amssymb} % assumes amsmath package installed
\usepackage{algorithm, algpseudocode} %algorithmic}
\usepackage{geometry}
\usepackage{bm}

\usepackage{graphicx}
\usepackage[caption=false,font=footnotesize]{subfig}
\usepackage{balance}

\usepackage[linkbordercolor={1 1 1},citebordercolor={1 1
  1},urlbordercolor={0.0 0.0
  0.0},urlcolor=blue,colorlinks=true,linkcolor=black,citecolor=black]{hyperref}
\usepackage{url}

\newcommand{\algo}{{\textsc{CoSpar}}}
\newcommand{\dimalgo}{{\textsc{LineCoSpar}}}

\newcommand{\newsec}[1]{\vspace{2mm} \noindent \textbf{#1.} }
\newcommand{\newsubsec}[1]{\vspace{2mm} \noindent \underline{#1.} }

\title{\vspace{10pt}\LARGE \bf
Human Preference-Based Learning for High-dimensional Optimization of Exoskeleton Walking Gaits
} 

\author{Maegan Tucker$^{1}$, Myra Cheng$^{2}$, Ellen Novoseller$^{2}$, Richard Cheng$^{1}$,\\ Yisong Yue$^{2}$, Joel W. Burdick$^{1,2}$, and Aaron D. Ames$^{1,2}$ % <-this % stops a space
\thanks{*This work was supported by NSF NRI award 1724464, NSF Graduate Research Fellowship No. DGE1745301, the Caltech Big Ideas Fund, and the ZEITLIN Fund.}% <-this % stops a space
\thanks{This work was conducted under IRB No. 16-0693.}% <-this % stops a space
\thanks{$^{1}$Authors are with the Department of Mechanical and Civil Engineering, California Institute of Technology, Pasadena, CA 91125}%
\thanks{$^{2}$Authors are with the Department of Computing and Mathematical Sciences, California Institute of Technology, Pasadena, CA 91125}%
}

\begin{document}

\maketitle
\thispagestyle{empty}
\pagestyle{empty}

%%%%%%%%%%%%%%%%%%%%%%%%%%%%%%%%%%%%%%%%%%%%%%%%%%%%%%%%%%%%%%%%%%%%%%%%%%%%%%%%
\begin{abstract}
Optimizing lower-body exoskeleton walking gaits for user comfort requires understanding users' preferences over a high-dimensional gait parameter space. However, existing preference-based learning methods have only explored low-dimensional domains due to computational limitations. To learn user preferences in high dimensions, this work presents \dimalgo, a human-in-the-loop preference-based framework that enables optimization over many parameters by iteratively exploring one-dimensional subspaces. Additionally, this work identifies gait attributes that characterize broader preferences across users. In simulations and human trials, we empirically verify that \dimalgo~is a sample-efficient approach for high-dimensional preference optimization. Our analysis of the experimental data reveals a correspondence between human preferences and objective measures of dynamicity, while also highlighting differences in the utility functions underlying individual users' gait preferences. This result has implications for exoskeleton gait synthesis, an active field with applications to clinical use and patient rehabilitation.

\end{abstract}

%%%%%%%%%%%%%%%%%%%%%%%%%%%%%%%%%%%%%%%%%%%%%%%%%%%%%%%%%%%%%%%%%%%%%%%%%%%%%%%%

% ****************** SECTIONS *******************
\section{INTRODUCTION}

Human-in-the-loop online learning techniques have demonstrated significant potential in human-robot interaction tasks \cite{bajcsy2017learning, christen2019guided, cremer2019model}, such as improving the performance of robotic assistive devices. In particular, online learning from human feedback can help to optimize walking gaits for lower-body exoskeletons \cite{tucker2019preference, zhang2017human, thatte2018method}, which are placed over existing limbs to assist mobility-impaired individuals. 

This work focuses on optimizing walking gaits for individual user comfort using the Atalante lower-body exoskeleton developed by Wandercraft. We use a pre-computed gait library, which generates gaits offline using optimization-based techniques from nonlinear dynamics and control \cite{agrawal2017first, harib2018feedback,gurriet2018towards}. Gaits are specified by parameters ranging from centers of pressure to step dimensions (step length, width, etc.).

Optimizing gait parameters for each exoskeleton user serves two purposes. First, it enables gait personalization to maximize each user's comfort. Second, the relationships among different users' preferences, in particular their optimal gaits, may provide insight into the properties of universally-preferred gaits. While some gait optimization approaches rely on numeric metrics such as the user's metabolic expenditure \cite{zhang2017human}, there are no metrics that have established correspondences with user comfort. For example, metabolic expenditure is not an appropriate metric as the exoskeleton does not require the user to expend effort towards walking. A quantitative understanding of human preferences could help generate new gait profiles to improve the existing gait library, which represents a small fraction of the rich space of human walking behaviors. This motivates optimizing over the high-dimensional space of exoskeleton gaits to characterize the utility functions governing users' gait preferences. 

We rely on users' pairwise preferences to learn exoskeleton gaits that optimize user comfort, as several studies have shown that for subjective human feedback, preferences are more reliable than numerical scores \cite{basu2017you, joachims2005accurately, chapelle2012large}.  While interactive preference learning methods have been applied to robotics \cite{tucker2019preference, thatte2018method}, existing online preference learning methods are restricted to low-dimensional domains due to computational limitations; for example, previous work on preference-based exoskeleton gait optimization either learns over at most two dimensions \cite{tucker2019preference} or utilizes domain knowledge to narrow the search space before performing online learning \cite{thatte2018method}.

 \begin{figure}[tb]
 \centering
 \includegraphics[width=0.95\linewidth]{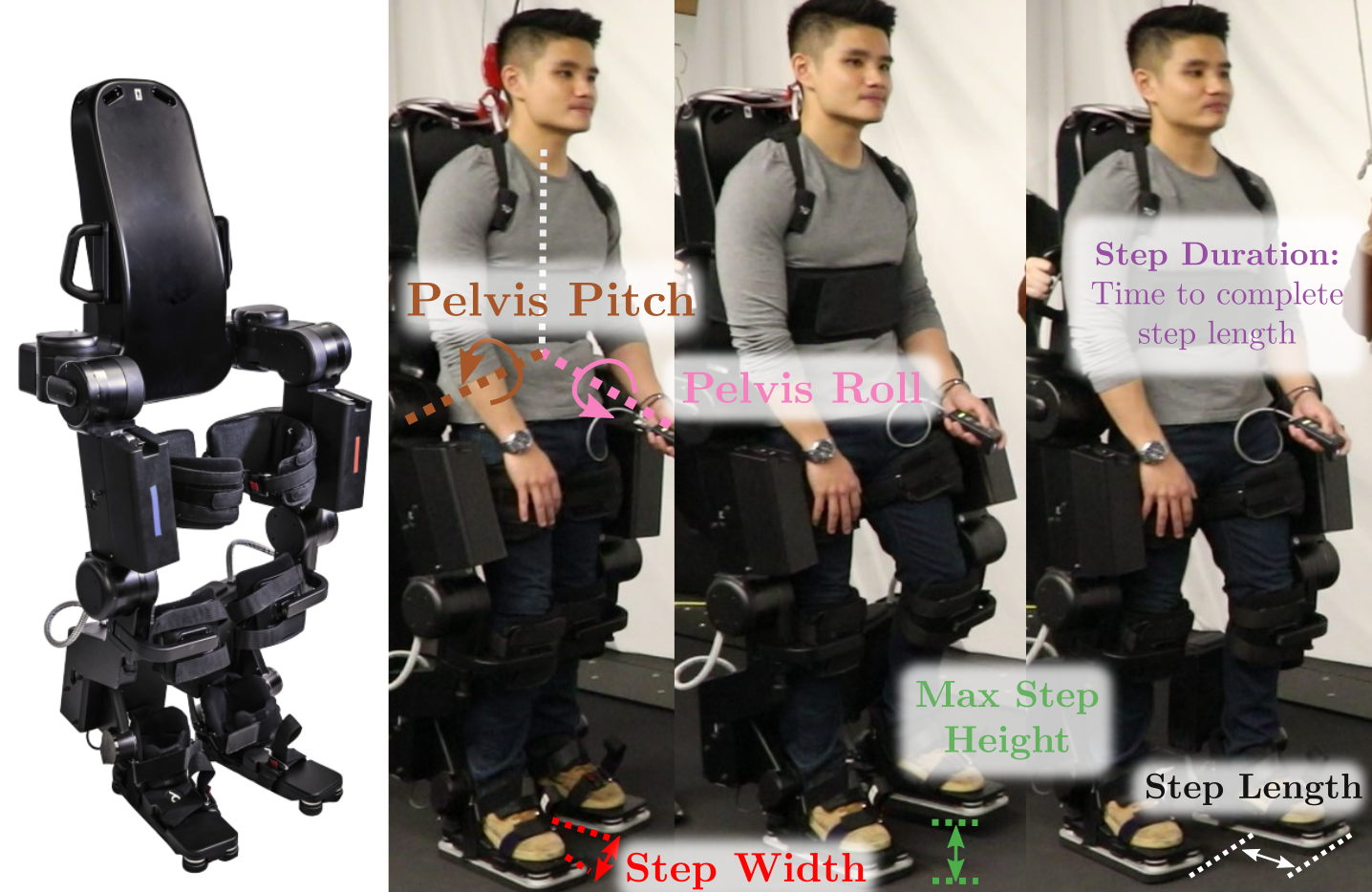}
%  \vspace{-0.02in}
 \caption{\textbf{Atalante Exoskeleton.} The exoskeleton has 12 actuated joints. The experiments explore six exoskeleton gait parameters: step length, step duration, step width, maximum step height, pelvis roll, and pelvis pitch.}
 \label{fig:Atalante}
 \end{figure}

% Sample efficiency in low-data settings is another major challenge; in clinical deployment of the exoskeleton with patients, data is expensive and difficult to obtain. Furthermore, preference feedback yields only one (possibly noisy) bit of data per pair of compared trials. To reduce sample complexity, we adopt a mixed-initiative approach, simultaneously querying the user for preferences and \textit{coactive feedback} \cite{shivaswamy2015coactive}, which allows the user to suggest improvements.

We present \dimalgo, a high-dimensional human preference-based learning approach that integrates existing techniques for preference learning \cite{tucker2019preference} and high-dimensional optimization \cite{kirschner2019adaptive} into a unified framework. \dimalgo~relies on preference feedback to iteratively explore one-dimensional subspaces.
We demonstrate in simulation that \dimalgo~exhibits sample-efficient convergence to user-preferred actions in high-dimensional spaces. The algorithm is then deployed experimentally to optimize exoskeleton walking over six gait parameters for six able-bodied subjects. 

Using the preferred gaits identified by \dimalgo~in the human experiments, we examine the connections among user-preferred gaits to understand what makes some gaits preferable to others. An analysis using the Zero Moment Point \cite{vukobratovic2004zero} reveals that users' preferences correspond to quantitative metrics of dynamicity. We observe that while most users' preferences are consistent with a metric that prioritizes dynamic stability, one user's preferences are explained by metrics that favor static stability. Based on this analysis, we suggest metrics that should be considered in the gait generation process, such that future exoskeleton gait designs can draw from regions of the gait trajectory space that prioritize user comfort.

% we examine the stability of gaits with respect to user preferences via the Zero Moment Point \cite{vukobratovic2004zero}. While \dimalgo~ identifies unique preferred gaits for each user, this analysis implies that there are also commonalities underlying user-preferred gaits, which help to explain why some gaits are preferable to others. We propose a gait stability metric that could be incorporated into the gait generation process, so that future exoskeleton gait design draws from regions of the gait trajectory space that prioritize user preference and comfort.
% TODO add details
% Since the  and find that exoskeleton users prefer gaits for which the center of mass exhibits less acceleration.

% This metric is a step toward quantifying and accounting for user comfort in exoskeleton gait generation beyond the existing gait library.             %|
\section{The Learning Algorithm}
\label{sec: learningAlgo}

The \dimalgo~algorithm (Alg. \ref{alg:coactive_self_sparring}) learns a Bayesian model over the user's preferences in a high-dimensional space. To learn from preferences, we adopt the dueling bandit setting \cite{sui2017multi, sui2018advancements, yue2012k}, in which the algorithm selects actions and receives relative preferences between them. The procedure, based on Thompson sampling, iterates through: 1) updating a Bayesian posterior over the actions' utilities given the data, 2) sampling utility functions from the posterior, 3) executing the actions that maximize the sampled utility functions, and 4) observing preferences among the executed actions. 

Drawing inspiration from the \textsc{LineBO} algorithm \cite{kirschner2019adaptive}, \dimalgo~exploits low-dimensional structure in the search space by sequentially considering one-dimensional subspaces from which to sample actions. This allows the algorithm to maintain its Bayesian preference relation function over a subset of the action space in each iteration. \dimalgo~builds upon \algo, which finds user-preferred parameters across one and two dimensions \cite{tucker2019preference}. Compared to \algo, \dimalgo~ learns the model posterior much more efficiently and can be scaled to higher dimensions.

% The \dimalgo~algorithm (Alg. \ref{alg:coactive_self_sparring}) builds upon \algo~\cite{tucker2019preference} by learning a Bayesian model over users' preferences in higher-dimensional spaces. Drawing inspiration from the \textsc{LineBO} algorithm \cite{kirschner2019adaptive}, \dimalgo~exploits low-dimensional structure in the search space by dividing the problem into a series of one-dimensional subproblems. This allows \dimalgo~to maintain its Bayesian preference relation function over only a subset of the action space in each iteration, reducing the computational complexity of learning the model posterior compared to \algo. 
This section provides background on existing approaches and then describes the \dimalgo~algorithm, including 1) defining the posterior updating procedure, 2) achieving high-dimensional learning, and 3) incorporating Thompson sampling and coactive feedback.

\subsection{Background}

\newsec{Preference-Based Learning}
We learn users' preferred exoskeleton gaits through their relative preferences, which are more reliable than subjective numerical feedback \cite{tucker2019preference, chapelle2012large, basu2017you, joachims2005accurately}. To maximize sample efficiency, we adopt the mixed-initiative approach of \algo \cite{tucker2019preference}, which learns from both pairwise preference and coactive feedback. In coactive learning \cite{shivaswamy2012online, shivaswamy2015coactive}, after each time the algorithm selects an action, the user identifies an improved action. Under both feedback types, the exoskeleton user tests various gaits to specify preferences and suggest gait modifications. \algo~effectively identifies user-preferred gait parameters across one and two dimensions. However, \algo~is intractable in larger action spaces, as it jointly maintains and samples from a posterior over every action, causing the computational complexity to increase exponentially in the action space dimension.

% To learn from preferences, we adopt the dueling bandit setting \cite{sui2017multi, sui2018advancements, yue2012k}, in which the algorithm selects actions and receives relative preferences between them. This builds upon \textsc{SelfSparring} \cite{sui2017multi}, a state-of-the-art Thompson sampling-based algorithm that iterates through: a) updating a Bayesian posterior over the actions' utilities given the data, b) sampling utility functions from the posterior, c) executing the actions that maximize the sampled utility functions, and d) observing preferences among the executed actions.
%\textsc{SelfSparring}, among other Bayesian posterior sampling algorithms, yields competitive theoretical guarantees and empirical performance \cite{sui2017multi, chapelle2011empirical, russo2014learning}.

\newsec{High-Dimensional Bayesian Optimization}
Bayesian optimization is a powerful approach for optimizing expensive-to-evaluate black-box functions. It maintains a model posterior over the unknown function, and cycles through a) using the posterior to acquire actions at which to query the function, b) querying the function, and c) updating the posterior using the obtained data. This procedure is challenging in high-dimensional search spaces due to the computational cost of the acquisition step \textit{(a)}, which often requires solving a non-convex optimization problem over the search space, and maintaining the posterior in the update step \textit{(c)}, which can require manipulating matrices that grow exponentially with the action space's dimension. Dimensionality reduction techniques are therefore an area of active interest. Solutions vary from optimizing variable subsets ($\textsc{DropoutBO}$) \cite{lidropoutbo} to projecting into lower-dimensional spaces ($\textsc{REMBO}$) \cite{wang2016rembo} to sequentially optimizing over one-dimensional subspaces ($\textsc{LineBO}$) \cite{kirschner2019adaptive}. We draw upon the approach 
of $\textsc{LineBO}$
because of its state-of-the-art performance in high-dimensional spaces. Furthermore, it is especially sample-efficient in spaces with underlying low-dimensional structure. In the case of exoskeleton walking, this low-dimensional structure may appear as linear relationships between two gait parameters in the user's utility function, i.e., users who prefer short step lengths also prefer short step durations.

\subsection{The \dimalgo~Algorithm}
\begin{algorithm}[tb]
\caption{\dimalgo}
\begin{small}
\begin{algorithmic}[1]
% $\mathcal{S}_t$ = sampled actions at iteration $t$, $\mathcal{L}_t$ = 1-dimensional subspace, $p_t$ = best point, 
\Procedure{\dimalgo}{Utility prior parameters; $m$ = granularity of discretization}
\State $\mathcal{D} = \emptyset$, $\mathcal{W} = \emptyset$ \Comment{$\mathcal{D}$: preference data, $\mathcal{W}$: actions in $\mathcal{D}$}
\State Set $\bm{p}_1$, $\bm{a}_0$ to uniformly-random actions
% \State Randomly initialize $\bm{p}_1$
\For{t = 1, 2,\dots, $T$}{}
 \State $\mathcal{L}_{t} = $ random line through $\bm{p}_{t}$, discretized via $m$
 \State $\mathcal{V}_t = \mathcal{L}_t \cup \mathcal{W}$  \Comment{Points over which to update posterior} 
 \State $(\bm{\mu}_{t}, \Sigma_{t}) = $ posterior over points in $\mathcal{V}_t$, given $\mathcal{D}$
\State Sample utility function $f_t \sim \mathcal{N}(\bm{\mu}_{t}, \Sigma_{t})$ \label{lin:sample_begin}
\State Execute action $\bm{a}_t = \text{argmax}_{\bm{a} \in \mathcal{V}_t} f_t(\bm{a})$ \label{lin:select_action}
\State Add pairwise preference between $\bm{a}_t$ and $\bm{a}_{t - 1}$ to $\mathcal{D}$
\State Add coactive feedback $\bm{a}_t^\prime$ to $\mathcal{D}$
\State Set $\mathcal{W} = \mathcal{W} \cup \{\bm{a}_t\} \cup \{\bm{a}_t^\prime\}$ \Comment{Update actions in $\mathcal{D}$}
\State Set $\bm{p}_{t + 1} = \text{argmax}_{\bm{a} \in \mathcal{V}_t} \mu_{t}(\bm{a})$
\EndFor
\EndProcedure
\end{algorithmic}
\end{small}
 \label{alg:coactive_self_sparring}
 \end{algorithm}

\newsec{Modeling Utilities Using Pairwise Preference Data} 
\label{subsec: modeling}
\dimalgo~uses pairwise comparisons to learn a Bayesian model posterior over the relative utilities of actions (i.e., gait parameter combination) to the user based upon the Gaussian process preference model in \cite{chu2005preference}.
We use Gaussian process learning, as it enables us to model a Bayesian posterior over a class of smooth, non-parametric functions.

Let $\mathcal{A} \subset \mathbb{R}^d$ be the set of possible actions. In iteration $t$ of the algorithm, we consider a subset of the actions $\mathcal{V}_t \subset \mathcal{A}$, with cardinality $V_t$ (we will define $\mathcal{V}_t$ later). We assume that each action $\bm{a} \in \mathcal{A}$ has a latent utility to the user, denoted as $f(\bm{a})$. Throughout the learning process, \dimalgo~stores a dataset of all user feedback, $\mathcal{D} = \{\bm{a}_{k_1} \succ \bm{a}_{k_2} \, | \, k = 1, \ldots, N\}$, consisting of $N$ preferences, where $\bm{a}_{k_1} \succ \bm{a}_{k_2}$ indicates that the user prefers action $\bm{a}_{k_1}$ to action $\bm{a}_{k_2}$. The preference data $\mathcal{D}$ is used to update the posterior utilities of the actions in $\mathcal{V}_t$. Defining $\bm{f} = [f(\bm{a}_{t_1}), f(\bm{a}_{t_2}), \ldots, f(\bm{a}_{t_{V_t}})]^T \in \mathbb{R}^{V_t}$, where $\bm{a}_{t_i}$ is the $i$\textsuperscript{th} action in $\mathcal{V}_t$, the utilities $\bm{f}$ have posterior:

\begin{equation}\label{eqn:posterior}
\mathcal{P}(\bm{f} | \mathcal{D}) \propto \mathcal{P}(\mathcal{D} | \bm{f})\mathcal{P}(\bm{f}).
\end{equation}
In each iteration $t$, we define a Gaussian process prior over the utilities $\bm{f}$ of actions in $\mathcal{V}_t$:
\begin{equation}
\mathcal{P}(\bm{f}) = \frac{1}{(2\pi)^{V_t/2} |\Sigma_t^{\text{pr}}|^{1/2}} \text{exp}\left(-\frac{1}{2} \bm{f}^T [\Sigma_t^{\text{pr}}]^{-1} \bm{f}\right),
\end{equation}
where $\Sigma_t^{\text{pr}} \in \mathbb{R}^{V_t \times V_t}$, $[\Sigma_t^{\text{pr}}]_{ij} = \mathcal{K}(\bm{a}_{t_i}, \bm{a}_{t_j})$, and $\mathcal{K}$ is a kernel. Our experiments use the squared exponential kernel. To compute the likelihood $\mathcal{P}(\mathcal{D} | \bm{f})$, we assume that the preferences may be corrupted by noise, such that:
\begin{equation}
\mathcal{P}(\bm{a}_{k_1} \succ \bm{a}_{k_2} | \bm{f}) = g\left(\frac{f(\bm{a}_{k_1}) - f(\bm{a}_{k_2})}{c}\right),
\end{equation}
where $g(\cdot) \in [0, 1]$ is a monotonically-increasing link function, and $c > 0$ is a hyperparameter indicating the degree of preference noise. 
% While \algo~assumes a Gaussian noise model, such that $g$ is the standard normal cumulative distribution function \cite{tucker2019preference, chu2005preference}, we found the heavier-tailed sigmoid distribution, $g_{\text{sig}}(x) := \sigma (x) = \frac{1}{1 + e^{-x}}$, to be a more effective link function.
While previous work uses the Gaussian cumulative distribution function for $g$ \cite{tucker2019preference, chu2005preference}, we empirically found that using the heavier-tailed sigmoid distribution, $g_{\text{sig}}(x) := \frac{1}{1 + e^{-x}}$, as the link function improves performance.
$g_{\text{sig}}(x)$ satisfies the convexity conditions for the Laplace approximation \cite{sigmoidconvex} and has been used to model preferences in other contexts \cite{sigmoidpref}. The full likelihood expression becomes:
\begin{equation}
  \mathcal{P}(\mathcal{D} | \bm{f}) = \prod_{k = 1}^N g_{\text{sig}}\left(\frac{f(\bm{a}_{k_1}) - f(\bm{a}_{k_2})}{c} \right).
\end{equation}

% Notice that we use the sigmoid function as the link function rather than the Gaussian

\noindent The posterior in \eqref{eqn:posterior} is estimated via the Laplace approximation as in \cite{chu2005preference}, yielding a multivariate Gaussian, $\mathcal{N}(\bm{\mu}_t, \Sigma_t)$.

\newsec{Sampling Approach for Higher Dimensions} Existing preference-based approaches optimize over the action space $\mathcal{A}$ by discretizing the entire space before beginning the learning process. This results in $m^d$ combinations from $m$ uniformly-spaced points (corresponding to actions) in each of the $d$ dimensions of $\mathcal{A}$. Thus, the cardinality of this set is $A := |\mathcal{A}| = m^d$; larger $m$ enables finer-grained search at a higher computational cost. The Bayesian preference model is updated over all $A$ points during each iteration.  This is intractable for higher $d$ since computing the posterior over $A$ points involves expensive matrix operations, such as inverting $\Sigma_t^{\text{pr}}, \Sigma_t \in \mathbb{R}^{A \times A}$. 

Inspired by \cite{kirschner2019adaptive}, \dimalgo~overcomes this intractability by iteratively considering one-dimensional subspaces (lines), rather than the full action space. In each iteration $t$, \dimalgo~selects uniformly-spaced points along a new random line $\mathcal{L}_t$ in the action space, which is determined by a uniformly-random direction and the action $\bm{p}_t$ that maximizes the posterior mean. Including $\bm{p}_t$ in the subspace encourages exploration of higher-utility areas. The posterior $\mathcal{P}(\mathcal{D} | \bm{f})$ is calculated over $\mathcal{V}_t := \mathcal{L}_t \,\cup\, \mathcal{W} $, where $\mathcal{W}$ is the set of actions for which $\mathcal{D}$ contains preference feedback. This approach reduces the model's covariance matrices $\Sigma_t^{\text{pr}}, \Sigma_t$ from size $A \times A$ to $V_t \times V_t$. Rather than growing exponentially in $d$, which is impractical for online learning, \dimalgo's complexity is constant in the dimension $d$ and linear in the number of iterations $T$. Since queries are expensive in many human-in-the-loop robotics settings, $T$ is typically low.

% Previous work \cite{tucker2019preference} discretizes each of the $d$ action space dimensions into $m$ bins, such that $\mathcal{A}$ is a finite action set with cardinality $A = |\mathcal{A}| = m^d$. Larger $m$ enables finer-grained search at a higher computational cost. This is intractable for higher $d$ since computing the posterior involves inverting $\Sigma_t^{\text{pr}}, \Sigma_t \in \mathbb{R}^{A \times A}$.
% as shown in Table \ref{table:matrixsize}.
% Inspired by \cite{kirschner2019adaptive}, \dimalgo~overcomes this intractibility by iteratively considering one-dimensional subspaces (i.e. lines), rather than considering the entire discretized action space at once. In each iteration $t$, \dimalgo~selects points along a new random line $\mathcal{L}_t$, which is determined by a uniformly-random direction and the action $\bm{p}_t$ that maximizes the posterior mean. Including $\bm{p}_t$ in the subspace encourages exploration of higher-utility areas. The posterior $\mathcal{P}(\mathcal{D} | \bm{f})$ is calculated over $\mathcal{V}_t = \mathcal{L}_t \,\cup\, \mathcal{W} $, where $\mathcal{W}$ is the set of actions over which $\mathcal{D}$ has preference feedback.
% 
% Critically, this approach shrinks the covariance matrix of the preference relation distribution, $\Sigma_t$, from size $A \times A$ to $V_t \times V_t$. Thus, the algorithm's complexity is constant in the dimension $d$ and linear in the iteration $t$, whereas the baseline algorithm's complexity grows exponentially in $d$ (Fig. \ref{fig:time}).

\begin{figure}
 \centering
 \includegraphics[width =\linewidth]{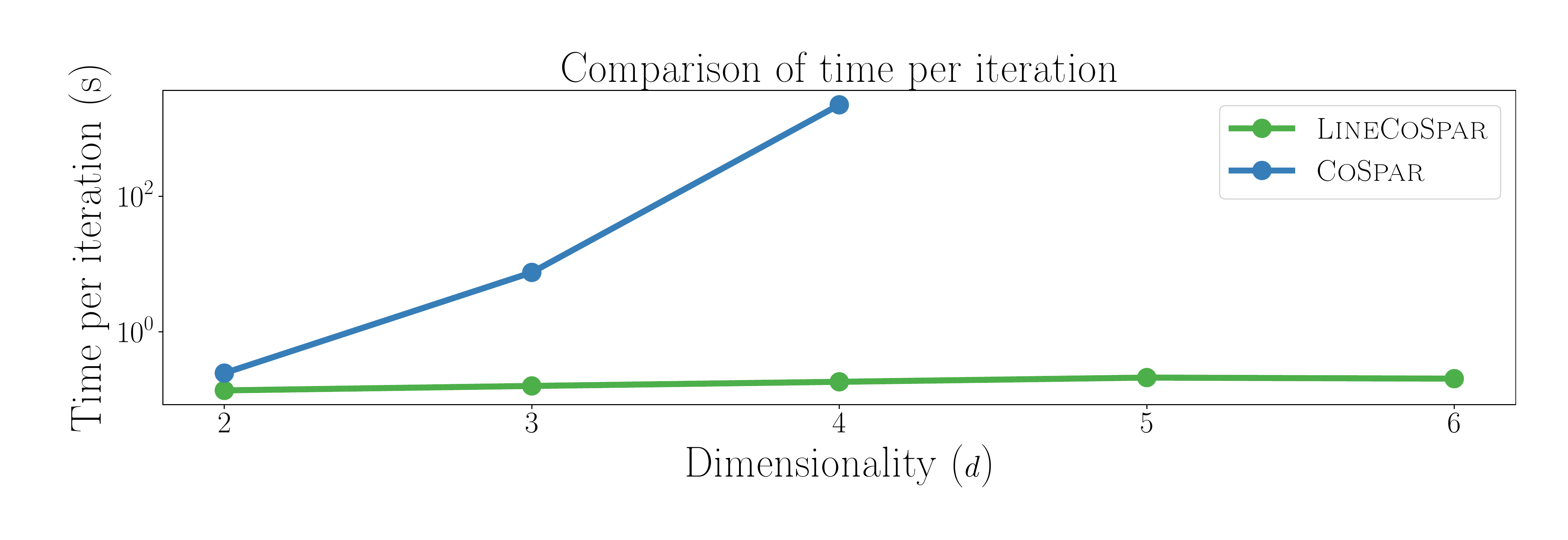}
 \vspace{-0.4in}
 \caption{\textbf{Curse of dimensionality for \algo.} Average time per iteration of \algo~vs. \dimalgo. The y-axis is on a logarithmic scale. For \dimalgo, the time is roughly constant in the number of dimensions $d$, while the runtime of \algo~increases exponentially. For $d = 4$, the duration of a \algo~iteration is inconvenient in the human-in-the-loop learning setting, and for $d\geq5$, it is intractable.}
 \vspace{-0.1in}
 \label{fig:time}
\end{figure}
\newsec{Posterior Sampling Framework}
Utilities are learned using the \textsc{SelfSparring} \cite{sui2017multi} approach to Thompson sampling detailed above. Specifically, in each iteration, we calculate the posterior of the utilities $\bm{f}$ over the points in $\mathcal{V}_t = \mathcal{L}_t \,\cup\, \mathcal{W}$, obtaining the posterior $\mathcal{N}(\bm{\mu}_t, \Sigma_t)$ over $\mathcal{V}_t$. The algorithm then samples a utility function $f_t$ from the posterior, which assigns a utility to each action in $\mathcal{V}_t$. Next, \dimalgo~executes the action $\bm{a}_t$ that maximizes $f$, $\bm{a}_t = \text{argmax}_{\bm{a} \in \mathcal{V}_t} f(\bm{a})$. The user provides a preference (or indicates indifference, i.e. ``no preference") between $\bm{a}_t$ and the preceding action $\bm{a}_{t - 1}$.

In addition, for each executed action $\bm{a}_t$, the user can provide coactive feedback, specifying the dimension, direction (higher or lower), and degree in which to change $\bm{a}_t$. The user's suggested action $\bm{a}_t^\prime$ is added to $\mathcal{W}$, and the feedback is added to $\mathcal{D}$ as $\bm{a}_t^\prime \succ \bm{a}_t$. In each iteration, preference and coactive feedback each add at most one action to $\mathcal{W}$. Thus, in iteration $t$, $\mathcal{V}_t$ contains at most $m + 2(t - 1)$ actions, and so its size is independent of the dimensionality $d$. In the subsequent analysis, $\bm{a}_{\text{max}}$ is defined as the action maximizing the final posterior mean after $T$ iterations, i.e., $\bm{a}_{\text{max}} := \text{argmax}_{\bm{a} \in \mathcal{V}_t} \mu_{T + 1}(\bm{a})$. 

% Note that \dimalgo~can be generalized to sample multiple actions per iteration, or to assume that the user can accurately remember (and make comparisons between) trials across non-consecutive iterations. Such extensions are considered in the definition of \algo~\cite{tucker2019preference}. 

% Analogously, $\bm{a}_{\text{min}} := \text{argmin}_{\bm{a} \in \mathcal{V}_t} \mu_{T + 1}(\bm{a})$.      %|
\section{Performance of \dimalgo}\label{sec:experiments}

\subsection{Simulation Results}
We validate the performance of \dimalgo~in simulation using both standard Bayesian optimization benchmarks and randomly-generated polynomials.\footnote{The code is at \url{https://github.com/myracheng/linecospar}. All experiments use the squared exponential kernel with lengthscale $0.15$ in every dimension, signal variance $1\mathrm{e}{-4}$, noise variance $1\mathrm{e}{-5}$, and preference noise 0.005.} The simulations show that \dimalgo~is sample-efficient, converges to sampling higher-valued actions, and learns a preference relation function such that actions with higher objective values have high posterior utilities.

\newsec{Standard Bayesian Optimization Benchmarks}
We evaluated the performance of \dimalgo~on the standard Hartmann3 (\textsc{H3}) and Hartmann6 (\textsc{H6}) benchmarks (3 and 6 dimensions, respectively). We do not compare \dimalgo~to other optimization methods because there are no other preference-based Gaussian process methods that are tractable in high dimensions. As discussed in Section \ref{subsec: modeling}, we focus on Gaussian process methods because they model smooth, non-parametric utility functions. We validate \dimalgo~with noiseless preferences and then demonstrate its robustness to noisy user preferences. Preferences are generated in simulation by comparing objective function values.

\begin{figure}
 \centering
 \includegraphics[width=\linewidth]{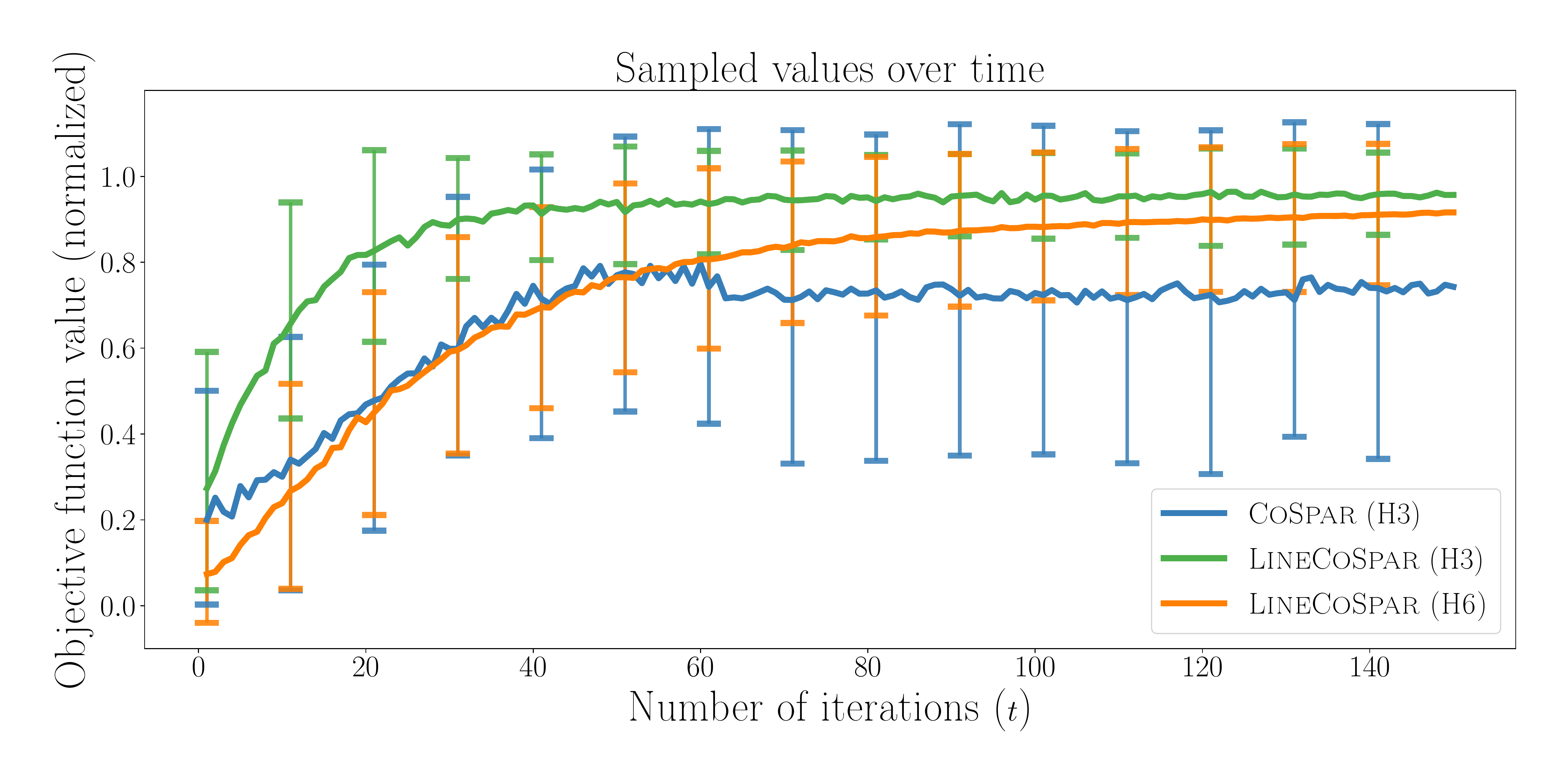}
 \vspace{-0.4in}
 \caption{\textbf{Convergence to higher values on standard benchmarks.} Mean objective value $\pm$ SD using H3 and H6, averaged over 100 runs. The sampled actions converge to higher objective values at a faster rate with \dimalgo, which has an improved sampling approach and link function. It is intractable to run \algo~on a 6-dimensional space.}
%  \vspace{-0.1in}
 \label{fig:336}
\end{figure}

Under ideal preference feedback, $\bm{a}_{k_1} \succ \bm{a}_{k_2}$ if $f(\bm{a}_{k_1}) > f(\bm{a}_{k_2})$. The true objective values $f$ are invisible to the algorithm, which observes only the preference dataset $\mathcal{D}$. Compared to \algo, \dimalgo~ converges to sampling actions with higher objective values at a faster rate (Fig. \ref{fig:336}). Thus, \dimalgo~not only enables higher-dimensional optimization, but also improves speed and accuracy of learning.

\begin{figure}
 \centering
 \includegraphics[width =\linewidth]{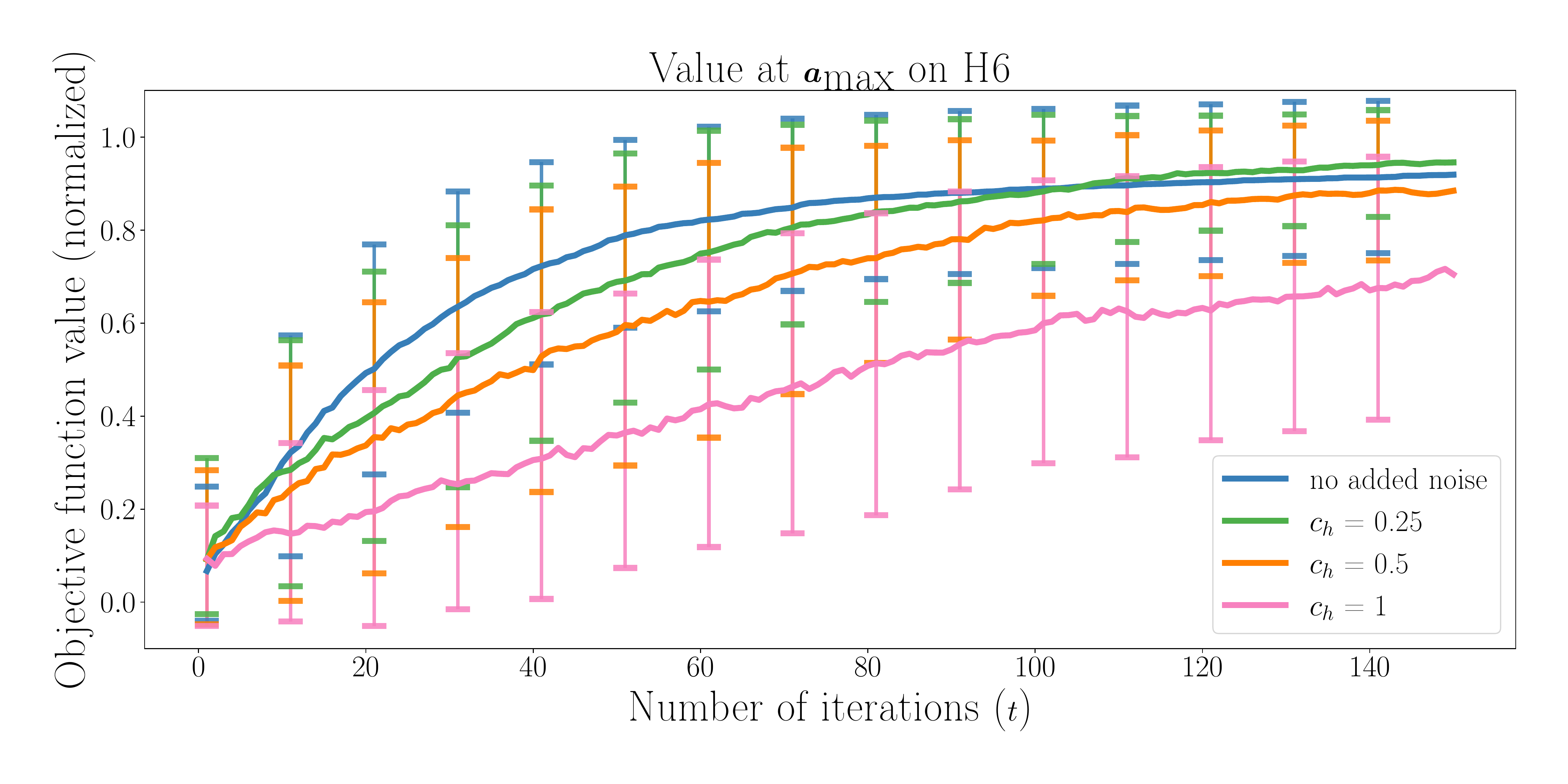}
\vspace{-0.4in}
 \caption{\textbf{Robustness to noisy preferences.} Mean objective value $\pm$ SD of the action $\bm{a}_{\text{max}}$ with the highest posterior utility. This is averaged over 100 runs using \dimalgo~on \textsc{H6} with varying preference noise, as quantified by $c_h$. Higher performance correlates with less noise (lower $c_h$). The algorithm is robust to noise to a certain degree ($c_h \leq 0.5$).}
%  \vspace{-0.1in}
 \label{fig:noise}
\end{figure}

 \begin{figure}[t]
 \centering
 \includegraphics[width=\linewidth]{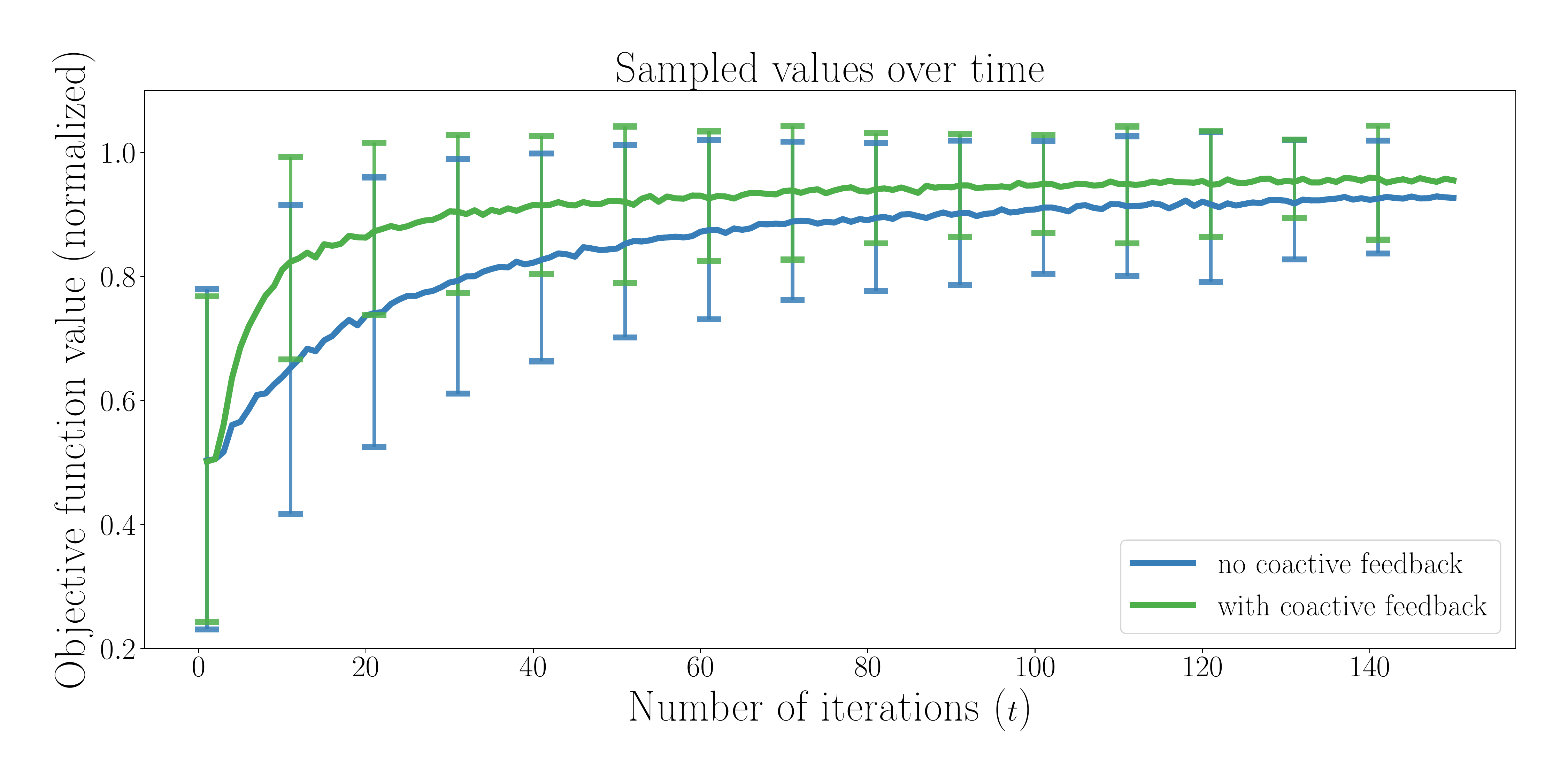}
 \vspace{-0.4in}
 \caption{\textbf{Coactive feedback improves convergence.} Mean objective value $\pm$ SD of the sampled actions using random functions. This is averaged over 1000 runs using \dimalgo~on 100 randomly-generated six-dimensional functions ($d=6$). The sampled actions converge to high objective values in relatively few iterations, and coactive feedback accelerates this process.}
%  \vspace{-0.1in}
 \label{fig:polysample}
 \end{figure}
 
  \begin{figure*}[t]
 \centering
 \includegraphics[width=\textwidth]{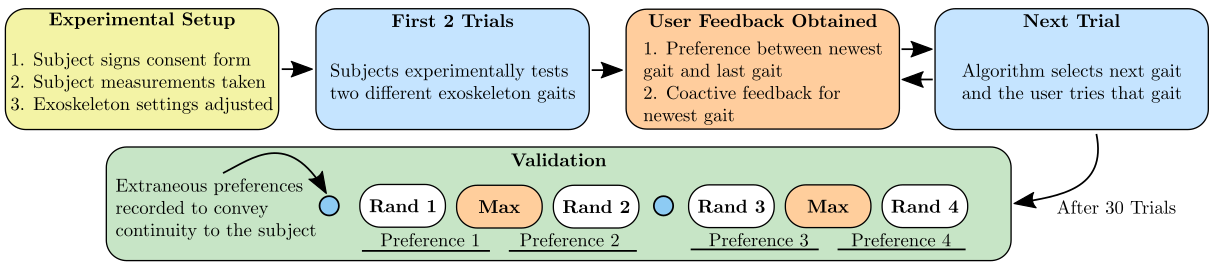}
 \vspace{-0.25in}
 \caption{\textbf{Experimental Procedure.} After setup of the subject-exoskeleton system, subjects were queried for preferences between all consecutive pairs of gaits, along with coactive feedback, in 30 gait trials (for a total of at most 29 pairwise preferences and 30 pieces of coactive feedback). After these 30 trials, the subject unknowingly entered the validation portion of the experiment, in which he/she validated the posterior-maximizing gait, $\bm{a}_{\text{max}}$, against four randomly-selected gaits.}
 \label{fig:procedure}
%   \vspace{-0.1in}
\end{figure*}

\begin{figure}[t]
 \centering
 \includegraphics[width=\linewidth]{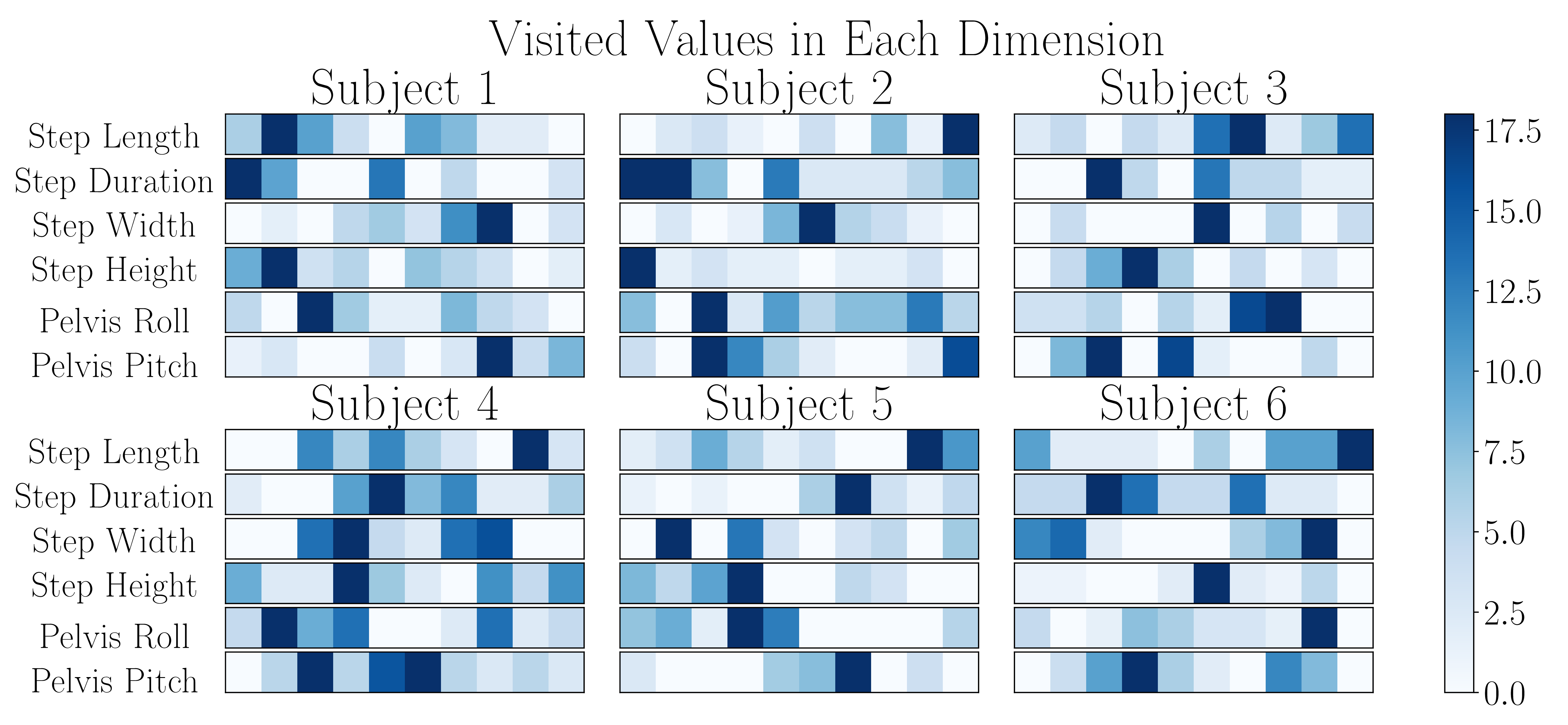} 
 \vspace{-0.25in}
 \caption{\textbf{Exploration vs. exploitation in human trials}. Each row depicts the distribution of a particular gait parameter's values across all gaits that the subject tested. Each dimension is discretized into 10 bins. Note that the algorithm explores different parts of the action space for each subject. These visitation frequencies exhibit a statistically-significant correlation with the posterior utilities across these regions (Pearson's p-value = 1.22e-10).}
 \label{fig:expexploration}
%   \vspace{-0.1in}
\end{figure}

% \begin{figure}[tb]
%  \centering
%  \includegraphics[width=\linewidth]{Figures/coactive.pdf}
%  \caption{\textbf{Coactive Feedback}. Each bar records the number of times that a subject provided coactive feedback regarding a certain gait parameter. Larger quantities of feedback suggest that the subject weighted the parameter more highly in determining the gaits' utilities. Perhaps, subjects also found it more intuitive to provide coactive feedback in some dimensions than others.
%  }
%  \label{fig:coac}
% \end{figure}
 
Since human preferences may be noisy, we tested the algorithm's robustness to noisy preference feedback. In simulation, this is modeled via $\mathcal P(\bm{a}_{k_1} \succ \bm{a}_{k_2}) = (1+e^{-\frac{s_k}{c_h}})^{-1}$, where $s_k = f(\bm{a}_{k_1}) - f(\bm{a}_{k_2})$ and $c_h$ is a hyperparameter for the noise level. As $c_h \rightarrow \infty$, the preferences approach uniform randomness (i.e. become noisier). Also, actions become less distinguishable when the distance between $f(\bm{a}_{k_1})$ and $f(\bm{a}_{k_2})$ decreases. This reflects human preference generation since it is more difficult to give consistent preferences between actions with similar utilities. By simulating noisy preferences, we demonstrate that \dimalgo~is robust to noisy feedback (see Fig. \ref{fig:noise}).

\newsec{Randomly-Generated Functions} We also tested \dimalgo~using randomly-generated $d$-dimensional polynomials (for $d$ = 6) as objective functions: $p(\bm{a}) = $ $\sum_{i=1}^{d}\alpha_i\sum_{j=1}^{d}\beta_j a_j$,
where $a_j$ denotes the $j$\textsuperscript{th} element of $\bm{a}$, and $\alpha_i,\beta_i, i \in \{1, \ldots, d\}$ are sampled independently from the uniform distribution $\mathcal{U}(-1,1)$.
% each $\alpha,\beta$ is randomly drawn from the uniform distribution $\mathcal{U}(-1,1).$
The dimensions' ranges and discretizations match those in the exoskeleton experiments, so that these simulations approximate the number of human trials needed to find optimal gaits.

Coactive feedback was simulated for each sampled action $\bm{a}_t$ by finding an action $\bm{a}_t^\prime$ with a higher objective value that differs from $\bm{a}_t$ along only one dimension. The action $\bm{a}_t^\prime$ is determined by randomly choosing a dimension in $\{1,\dots,d\}$ and direction (positive or negative), and taking a step from $\bm{a}_t$ along this vector. If the resulting action $\bm{a}_t^\prime$ has a higher objective value, it is added to the dataset $\mathcal{D}$ as $\bm{a}_t^\prime \succ \bm{a}_t$. This is a proxy for the human coactive feedback acquired in the exoskeleton experiments described below, in which the user can suggest a dimension and direction in which to modify an action to obtain an improved gait.

Fig. \ref{fig:polysample} displays \dimalgo's performance over 100 randomly-generated polynomials (10 repetitions each) with computation time shown in Fig. \ref{fig:time}. The results demonstrate that \dimalgo~samples high-valued actions within relatively few iterations ($\approx$ 20 with coactive feedback).

\subsection{Human Subject Experiments}
After the performance of \dimalgo~ was demonstrated in simulation, the algorithm was experimentally deployed on the lower-body exoskeleton Atalante (Fig. \ref{fig:Atalante}) to optimize six gait parameters for six able-bodied users (see Table \ref{table:expresults} for results and \cite{video} for a video).

\newsec{Atalante Exoskeleton}
Atalante (Fig. \ref{fig:Atalante}) \cite{harib2018feedback, duburcq2019online, gurriet2019towards}, developed by Wandercraft, has 12 actuated joints: three at each hip, one at each knee, and two in each ankle. \cite{gurriet2018towards} describes the device's mechanical components and control architecture in detail. Exoskeleton walking is achieved using pre-computed walking gaits, generated using the partial hybrid zero dynamics framework \cite{ames2014human} and a nonlinear constrained optimization process that utilizes direct collocation. The configuration space of the human-exoskeleton system is constructed as $q = (p,\phi, q_b) \in Q \in \mathbb{R}^{18}$, where $p \in \mathbb{R}^3$ and $\phi \in \mathbb{SO}^3$ denote the position and orientation of the exoskeleton floating base frame with respect to the world frame, and $q_b \in \mathbb{R}^{12}$ denotes the relative angles of the actuated joints. The generated gaits are realized on the exoskeleton using PD control at the joint level and a high-level controller adjusting joint targets based on state feedback. The controller is executed by an embedded computer unit running a real-time operating system. Gaits are sent to the exoskeleton over a wireless connection via a custom graphical user interface. 

\begin{table*}[tb]
\caption{Gait parameters optimizing \dimalgo's posterior mean ($\bm{a}_{\text{max}}$) for each able-bodied subject}
\label{table:expresults}
\vspace{-0.2in}
\begin{center}
\begin{tabular}{|c|c|c||c|c|c|c|c|c||c|}
\hline
\textbf{Subject} & \textbf{\begin{tabular}[c]{@{}c@{}}Height\\ (m)\end{tabular}} & \textbf{\begin{tabular}[c]{@{}c@{}}Mass\\ (kg)\end{tabular}} & \textbf{\begin{tabular}[c]{@{}c@{}}Step Length\\ (m)\end{tabular}} & \textbf{\begin{tabular}[c]{@{}c@{}}Step Duration\\ (s)\end{tabular}} & \textbf{\begin{tabular}[c]{@{}c@{}}Step Width\\ (m)\end{tabular}} & \textbf{\begin{tabular}[c]{@{}c@{}}Max Step \\ Height (m)\end{tabular}} & \textbf{\begin{tabular}[c]{@{}c@{}}Pelvis Roll\\ (deg)\end{tabular}} & \textbf{\begin{tabular}[c]{@{}c@{}}Pelvis Pitch\\ (deg)\end{tabular}} & \textbf{\begin{tabular}[c]{@{}c@{}}Validation\\ Accuracy (\%)\end{tabular}} \\ \hline
\textbf{1}  & 1.85          & 89.9          & 0.0835           & 0.943           & 0.278           & 0.0674           & 6.38            & 10.9           & \textbf{75}               \\ \hline
\textbf{2}  & 1.668          & 69.2          & 0.136           & 1.04            & 0.285           & 0.0679           & 6.41            & 12.4           & \textbf{100}               \\ \hline
\textbf{3}  & 1.635          & 51.2          & 0.137           & 0.922           & 0.279           & 0.0688           & 8.56            & 11.4           & \textbf{100}               \\ \hline
\textbf{4}  & 1.795          & 73.6          & 0.127           & 0.989           & 0.268           & 0.065            & 6.68            & 12.7           & \textbf{25}              \\ \hline
\textbf{5}  & 1.625          & 55.9          & 0.161           & 1.05            & 0.258           & 0.0689           & 7.32            & 13.2           & \textbf{100}               \\ \hline
\textbf{6}  & 1.66          & 65           & 0.177           & 1.11            & 0.256           & 0.0663           & 7.71            & 13.5           & \textbf{100}               \\ \hline
\end{tabular}
\end{center}
  \vspace{-0.1in}
\end{table*}

\newsec{Experimental Procedure}
\dimalgo~optimized exoskeleton gaits for six self-identified able-bodied subjects over six gait parameters (Fig. \ref{fig:Atalante}): step length, step duration, step width, maximum step height, pelvis roll, and pelvis pitch. These parameters were chosen from the pre-computed gait library because they are relatively intuitive for users to understand when giving coactive feedback. The parameter ranges, respectively, are: 0.08-0.18 meters, 0.85-1.15 seconds, 0.25-0.3 meters, 0.065-0.075 meters, 5.5-9.5 degrees, and 10.5-14.5 degrees. Fig. \ref{fig:procedure} illustrates the experimental procedure for testing and validating \dimalgo. 

All subjects were volunteers without prior exoskeleton exposure. For each subject, the testing procedure lasted approximately two hours, with one hour of setup and one hour of exoskeleton testing. The setup consisted of explaining the procedure (including how to provide preference and coactive feedback), measuring subject parameters, and adjusting the thigh and shank length of the exoskeleton to the subject. During the testing, the subjects had control over initiating and terminating each instance of exoskeleton walking and were instructed to try each walking gait until they felt comfortable giving a preference. The subjects could choose to test each gait multiple times to confirm their preference. They could also specify ``no preference'' between two gait trials, in which case no new information was added to the dataset $\mathcal{D}$.

After completing 30 trials (including trials with no preference, but not including voluntary gait repetitions), the subject began a set of ``validation'' trials; for consistency, the subject was not informed of the start of the validation phase. Validation consisted of six additional trials and yielded four pairwise preferences, each between the posterior-maximizing action $\bm{a}_{\text{max}}$ and a randomly-generated action. This validation step verifies that $\bm{a}_{\text{max}}$ is preferred over other parameter combinations across the search space.

\newsec{Gait Optimization Results}
Fig. \ref{fig:expexploration} shows that the \dimalgo~algorithm both explores across the gait parameter space and exploits regions with higher posterior utility. Over time, \dimalgo~ increasingly samples actions concentrated in regions of the search space that are preferred based on previous feedback. This results in a significant correlation between visitation frequencies and posterior utilities across these regions (Pearson's p-value = 1.22e-10).

For each subject, Table \ref{table:expresults} lists the parameters of the predicted optimal gaits, $\bm{a}_{\text{max}}$, identified by \dimalgo. Table \ref{table:expresults} also illustrates the results of the validation trials for each subject. These results show that $\bm{a}_{\text{max}}$ was predominantly preferred over the randomly-selected actions during validation. For four of the six subjects, all four validation preferences matched the posterior, while the other subjects matched three and one of the four preferences, respectively. %\textcolor{red}{Outliers may be due to noisy preference feedback that did not reflect the users' true utilities.}

% \begin{figure*}[t!]
%  \centering
%  \includegraphics[width=\linewidth]{correlations.pdf}
%  \caption{\textbf{Trends among actions maximizing the preference relation posterior.} Ten strongest correlations (based on $r^2$ values and associated p-values) among gait parameters and user features. These correlations suggest various relationships. The pelvis pitch, step duration, and step length parameters appear to have stronger trends than pelvis roll, step height, and step width. Identifying these correlations may help to illuminate characteristics of preferred gaits.}
%  \label{fig:corr}
% \end{figure*}                  %|
\section{Analysis of Preference Feedback and Implications for Gait Synthesis}
In addition to optimizing exoskeleton walking gaits for individual users, we aim to understand the utility functions underlying human preferences and apply this knowledge towards improving gait synthesis. As discussed in \cite{tucker2019preference}, exoskeleton gaits are generated using the partial hybrid zero dynamics framework, which is formulated by the following nonlinear optimization problem \cite{harib2018feedback}:
\begin{align*}
    \alpha^{*} = \mathop{\mathrm{argmin}}\limits_{\alpha} \quad &\mathcal{J}(\alpha)\\
    \textrm{s.t.} \quad &\Delta(\mathcal{S}\cap \mathcal{PZ}_{\alpha}) \subset \mathcal{PZ}_{\alpha} \\
    &\mathcal{W}_ix \leq b_i \\
    &\dot{\eta_{\alpha}} = A_{cl} \eta_{\alpha},
\end{align*}
where $\alpha$ are coefficients of B\'ezier polynomials that yield impact-invariant periodic orbits, $\mathcal{J}(\alpha)$ is a user-determined cost, $\Delta(\mathcal{S}\cap \mathcal{PZ}_{\alpha}) \subset \mathcal{PZ}_{\alpha}$ is the impact invariance condition, $\mathcal{W}_ix \leq b_i$ are other physical constraints, and $\dot{\eta_{\alpha}} = A_{cl} \eta_{\alpha}$ is the output dynamics condition. For more details on these constraints, refer to \cite{ames2014human}.

The cost function $\mathcal{J}(\alpha)$ largely influences the behavior of the walking gaits that it generates; however, the user's cost function $\mathcal{J}_{\text{human}}$ underlying her preferences is poorly-understood. This section aims to describe the relationship between gaits and user preferences through the underlying cost function $\mathcal{J}_{\text{human}}$, so that future gait synthesis can be streamlined towards user-preferred walking. Thus, we aim to identify key terms in $\mathcal{J}_{\text{human}}$ that numerically account for the preferences captured by \dimalgo.

All walking gaits on the exoskeleton are flat-footed. Thus, by analyzing the center of mass (CoM) and center of pressure (CoP), we can treat the patient-exoskeleton system as a Linear Inverted Pendulum Model (LIPM). This allows us to analyze the underlying utility function $\mathcal{J}_{\text{human}}$ using the cost structure from \cite{stephens2010push}. We first introduce Zero Moment Point (ZMP) and LIPM, and then discuss correspondences between metrics of dynamic stability and user comfort. 

\begin{figure}[tb]
  \includegraphics[width = \linewidth]{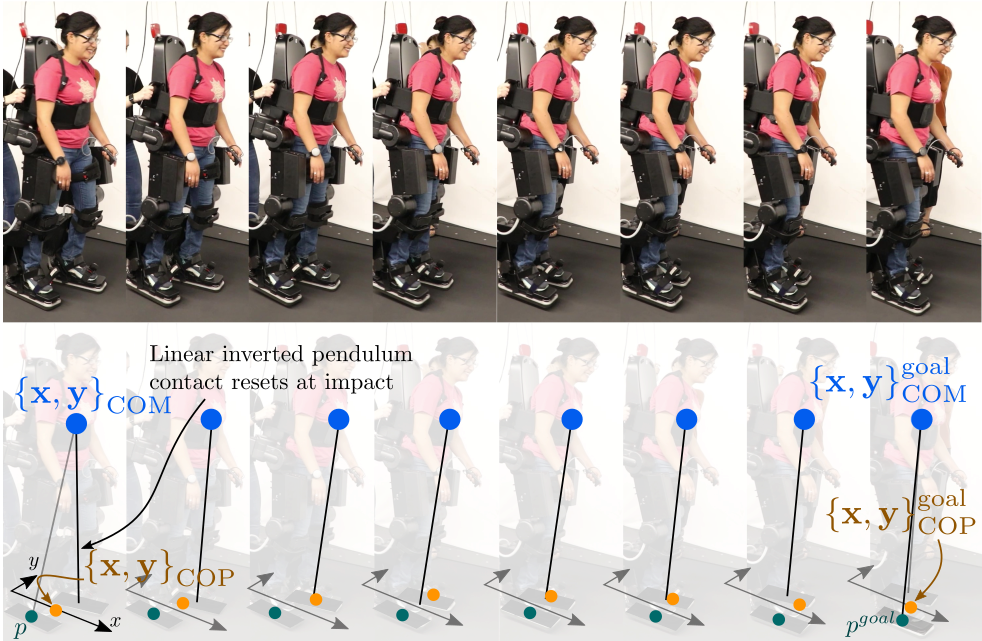}
  \vspace{-0.2in}
  \caption{Illustration of a single step with the overlayed LIPM model.}
  \label{fig:zmp}
  \vspace{-0.1in}
\end{figure}

\newsubsec{Zero Moment Point}
The Zero Moment Point (ZMP) is a widely-used notion of stability for bipedal robots that is defined as \textit{the point on the ground at which the net moment of the inertial forces and the gravity forces has no component along the horizontal axes} \cite{vukobratovic2004zero}. When the ZMP exists outside of the ``support polygon,'' i.e. the convex hull of the stance foot (or stance feet in the double-support domain), the robot experiences foot roll.

\newsubsec{Static and Dynamic Stability} For a full discussion, refer to pg. 7 of \cite{westervelt2018feedback}. In general, static stability is the condition in which the CoM and CoP never leave the support polygon. In contrast, quasi-static stability relaxes this condition on the CoM and only requires that the CoP remains inside the support polygon. For dynamic stability, the CoP lies on the boundary of the support polygon for a portion of the gait.

\begin{figure*}[tb]
    \centering
    \includegraphics[width=\linewidth]{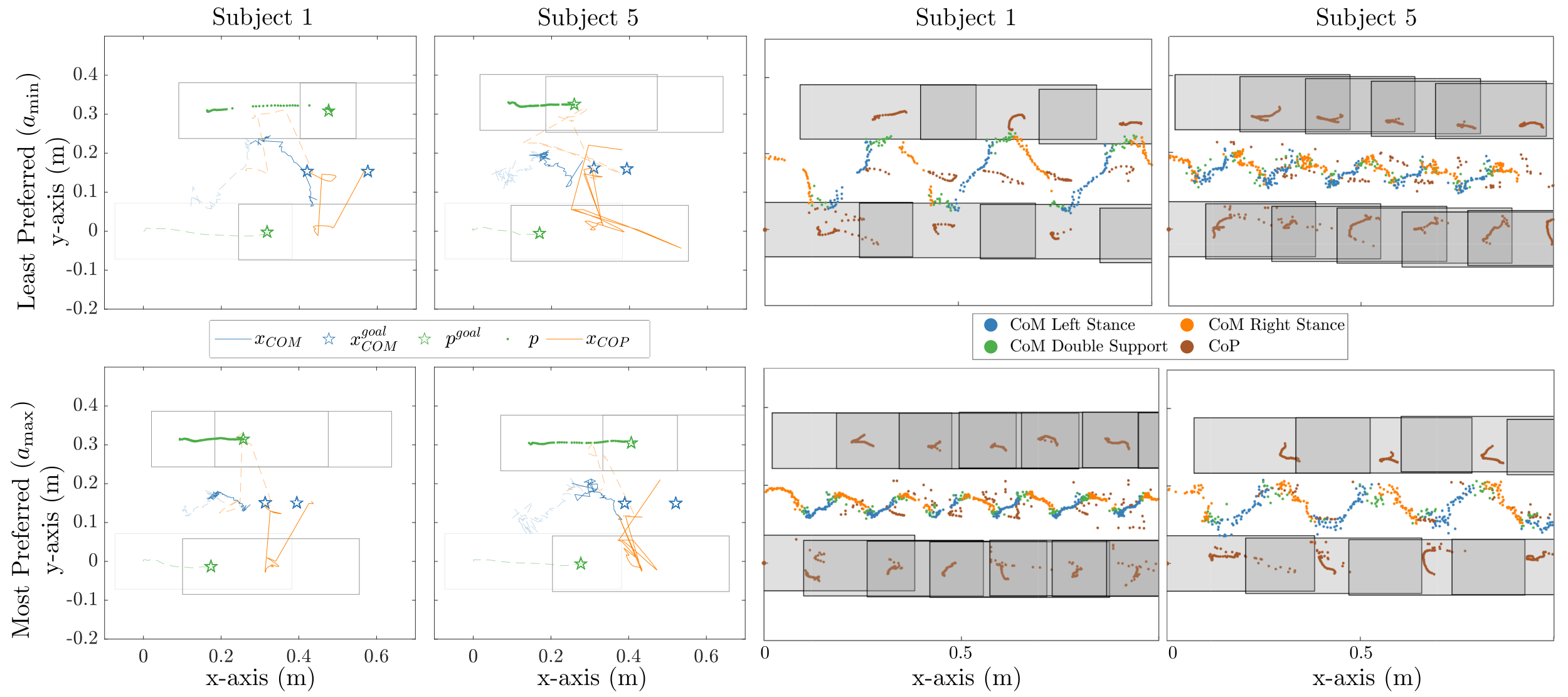}
    \vspace{-0.3in}
    \caption{\textbf{Comparison of Preferences}. This figure illustrates the trade-off between more and less dynamically-stable gaits as well as the contrasting preferences among different subjects. While all of the exoskeleton gaits are dynamically stable, both the least preferred gait $(a_{\text{min}})$ of subject 1 and the most preferred gait $(a_{\text{max}})$ of subject 5 exhibit behavior closer to statically-stable gaits. Subject 1 preferred dynamic gaits with a large difference between $x_{\text{CoP}}$ and $x_{\text{CoM}}$; in contrast, subject 5 preferred gaits in which $x_{\text{CoP}}$ closely followed the center of mass. Rectangles represent the exoskeleton's feet.}
    \label{fig:comparison}
\end{figure*}

\newsubsec{Linear Inverted Pendulum Model (LIPM)}  The LIPM is a low-dimensional dynamical system for reduced-order gait generation. The LIPM model assumes constant height of the center of mass, as well as zero angular momentum. The dynamics of the LIPM \cite{kajita2003biped} are:
\begin{align*}
  m\ddot{x}_{\text{CoM}} &= \frac{mg}{z_0}(x_{\text{CoM}}-x_{\text{CoP}}), \\
  m\ddot{y}_{\text{CoM}} &= \frac{mg}{z_0}(y_{\text{CoM}}-y_{\text{CoP}}),
\end{align*}
where $\{x,y\}_{\text{CoM}}$ are the $x$ and $y$ positions  of the CoM at constant height $z_0$, and $\{x,y\}_{\text{CoP}}$ denote the $x$ and $y$ positions of the CoP. For planar horizontal ground walking, the ZMP is mathematically equivalent to the CoP. The CoP was experimentally obtained using the four 3-axis force sensors on the bottom of the exoskeleton's feet. 

% The MPC QP formulation for generating gaits for the LIPM is similar to the PHZD formulation:
% \begin{align*}
    % u^{*} = \mathop{\mathrm{argmin}}\limits_{u} \quad \mathcal{J}(w,x,u) \hspace{2cm}&\\
%     \textrm{s.t.} \quad X_{t+1} = AX_t + BU_t &\\
%     \{x,y\}_{\text{CoP}}^\text{min} \leq \{x,y\}_{\text{CoP}} \leq \{x,y\}_{CoP}^{max}&
% \end{align*}
% where the first constraint is the linearized discrete dynamics, the second constraint is the ZMP constraints, and the cost function is any quadratic cost function. For more details on this formulation refer to \cite{stephens2010push}. 

\newsec{Fitting the LIPM Cost Function to User Preferences}
Since flat-foot level-ground walking is well captured by the LIPM model, the cost function used in the LIPM to generate desirable walking behavior may explain the users' utility functions underlying their exoskeleton gait preferences. As defined in \cite{stephens2010push}, the LIPM cost function is:
\begin{align*}
  & & \mathcal{J}_{\text{LIPM}} = w_1 ||x^{goal}_{\text{CoM}} - x_{\text{CoM}}||^2 + w_2 ||\dot{x}_{\text{CoM}}||^2 &+ \\
    & & w_3 ||\dot{x}_{\text{CoP}} ||^2 + w_4 ||p_x^{\text{goal}} - p_x||^2 &+\\
    & & w_1 ||y^{\text{goal}}_{\text{CoM}} - y_{\text{CoM}}||^2 + w_2 ||\dot{y}_{\text{CoM}}||^2 &+\\
    & & w_3 ||\dot{y}_{\text{CoP}}||^2 + w_4 ||p_y^{\text{goal}} - p_y||^2, &
\end{align*}
where $\{x,y\}^{\text{goal}}_{\text{CoM}}$ denotes the CoM goal position in the $x$ and the $y$ directions, $\{\dot{x},\dot{y}\}_{\text{CoP}}$ denotes the velocity of the CoP in the $x$ and $y$ directions, $\{\dot{x},\dot{y}\}_{\text{CoM}}$ is the velocity of the CoM, $p_{\{x,y\}}^{\text{goal}}$ denotes the next stance foot position in the $x$ and $y$ directions, and $p_{\{x,y\}}$ denotes the $x$ and $y$ positions of the swing foot (Fig. \ref{fig:zmp}).

We hypothesize that $\mathcal{J}_{\text{human}}(w)$ can be captured as a function of the weights $w := \{w_i\}, i \in \{1, \ldots, 4\}$.
Therefore, we fit the weights $w$ of $\mathcal{J}_{\text{LIPM}}$ to the validation-stage preference data, i.e., the preferences between the most-preferred gaits (gaits with parameters $\bm{a}_{\text{max}}$) and each of the random gaits presented during the validation phase\footnote{\label{analysis_code}Cost function fitting and CoP/CoM plotting code can be found at: \newline \url{https://github.com/myracheng/linecospar/tree/master/gaitAnalysis}}. The weights $w$ were optimized via the quadratic program: 
\begin{align*}
w* &= \mathop{\mathrm{argmin}}\limits_{w}\  ||w|| \quad \\
 \text{s.t.} \, &\begin{bmatrix} \delta_{1}^{(1)} & \delta_2^{(1)} & \delta_3^{(1)} & \delta_4^{(1)} \\
  & & \vdots & & \\
  \delta_{1}^{(n)} & \delta_2^{(n)} & \delta_3^{(n)} & \delta_4^{(n)} 
  \end{bmatrix} 
  \begin{bmatrix} w_1 \\ w_2 \\ w_3 \\ w_4 \end{bmatrix} < 0,
\end{align*}
where $n$ denotes the number of pairwise preferences, and:
\begin{align*}
  \begin{matrix}   
  \delta_{i} = \left( ||x^{\text{pref}}_{(i,x)}||^2 + ||x^{\text{pref}}_{(i,y)}||^2 \right) - \left( ||x^{\text{not pref}}_{(i,x)}||^2 + ||x^{\text{not pref}}_{(i,y)}||^2 \right) \\
  \end{matrix}  \\
  \begin{array}{ll}     
  x_{(1,x)} = x^{goal}_{\text{CoM}} - x_{\text{CoM}} \quad &
  x_{(1,y)} = y^{goal}_{\text{CoM}} - y_{\text{CoM}} \\
  x_{(2,x)} = \dot{x}_{\text{CoM}} & 
  x_{(2,y)} = \dot{y}_{\text{CoM}} \\
  x_{(3,x)} = \dot{x}_{\text{CoP}} &
  x_{(3,y)} = \dot{y}_{\text{CoP}} \\
  x_{(4,x)} = p_x^{goal} - p_{x} &
  x_{(4,y)} = p_y^{\text{goal}} - p_{y}.
  \end{array}    
\end{align*}
We use subject-wise holdout (leave-one-out) cross-validation across the subjects to verify the reliability of the fit. The average weights across all six holdout fits are: $w_1 = -0.1266$, $w_2 = 0.1363$, $w_3 = -0.0944$, and $w_4 = 1.0662$.

We quantify the predictive power of each fitted cost function on the users' utility functions using the rank consistency between the cost function values and the preference data. Table \ref{table:fitting} shows the predictive power of $\mathcal{J}_{\text{LIPM}}$ on the preferences, as well as the predictive power of two other cost functions, $\mathcal{J}_{\text{static}}$ and $\mathcal{J}_{\text{dynamic}}$, respectively defined as:
\begin{gather*}
  \mathcal{J}_{\text{static}} = ||\{x,y\}_{\text{CoM}} - \{x,y\}_{\text{CoP}}||^2, \\
    \mathcal{J}_{\text{dynamic}} = ||p_{\{x,y\}}^{\text{goal}} - p_{\{x,y\}}||^2.
\end{gather*}
These two metrics are directly opposed: while $\mathcal{J}_{\text{dynamic}}$ is the term from $\mathcal{J}_{\text{LIPM}}$ that promotes dynamic stability, $\mathcal{J}_{\text{static}}$ penalizes dynamic stability in favor of static stability. This is because in the LIPM dynamics, the acceleration of $\{x,y\}_{\text{CoM}}$ approaches zero as $\mathcal{J}_{\text{static}}$ approaches zero. We find that $\mathcal{J}_{\text{LIPM}}$ and $ \mathcal{J}_{\text{dynamic}}$ capture the preferences of five of the six subjects, while $\mathcal{J}_{\text{static}}$ completely predicts the preferences of the single outlier, subject 5.
%  To understand the underlying utility function of the outlier subject, we explored other possible cost functions and discovered that the following aligns best with the outlier user's preferences: 
% This term penalizes dynamic stability in favor of static stability as shown by the dynamics of the LIPM (the acceleration of $\{x,y\}_{\text{CoM}}$ goes to zero as this term goes to zero). Interestingly, this metric directly opposes the dynamic stability metric in $\mathcal{J}_{\text{LIPM}}$:

Fig. \ref{fig:comparison} further illustrates this difference. The largest discrepancy between $\mathcal{J}_{\text{dynamic}}$ and $\mathcal{J}_{\text{static}}$ is that of subject 1 and subject 5. The preferences of subject 1 align with dynamic stability, while the preferences of subject 5 align with static stability. The diametric opposition between the cost function terms predicting these users' preferences reflects inconsistencies across users' gait utility functions. This suggests that there is most likely no single metric that entirely captures all users' underlying utilities. Thus, it is important to generate a variety of gaits that satisfy the cost functions reflecting different users' preferences.

\begin{table}[tb]
\caption{Predictive power of cost functions on user preferences}
\label{table:fitting}
%\vspace{-0.2in}
\begin{tabular}{|c|c|c|c|c|c|c|}
\hline
{\textbf{Cost Function}}        & \multicolumn{6}{c|}{\textbf{\begin{tabular}[c]{@{}c@{}}Correctly predicted preferences per subject (\%)\end{tabular}}} \\ \cline{2-7} 
                                               & \textbf{1}          & \textbf{2}          & \textbf{3}          & \textbf{4}         & \textbf{5}         & \textbf{6}         \\ \hline
\textbf{$\mathcal{J}_{\text{LIPM}}$ (holdout)} & 75                  & 100                 & 62.5                & 75                 & 12.5               & 87.5                \\ \hline
\textbf{$\mathcal{J}_{\text{LIPM}}$}           & 75                  & 87.5                & 62.5                & 75                 & 62.5               & 100                \\ \hline
\textbf{$\mathcal{J}_{\text{dynamic}}$}        & 100                 & 100                 & 50                  & 75                 & 12.5               & 37.5               \\ \hline
\textbf{$\mathcal{J}_{\text{static}}$}         & 50                  & 75                  & 37.5                & 50                 & 100                & 75                 \\ \hline
\end{tabular}
\end{table}  

% is suggests that one quantitative metric cannot entirely capture all users' preferences; on the contrary, users may have utility functions that are diametrically opposed.                 %|
\section{CONCLUSION}

This work presents two main contributions: 1) the \dimalgo~algorithm to efficiently learn personalized, user-preferred gaits in high dimensions, and 2) an approach for understanding the mechanisms dictating
individual users' gait preferences.

\dimalgo~identifies preferred actions in high dimensions, both in simulation and in experiments with six able-bodied subjects using the Atalante lower-body exoskeleton. We then examine the experimentally-obtained gait preferences to gain insight into the utility functions underlying users' gait preferences. We identify opposing measures of dynamicity that have predictive power for different users' preferences, implying that each user consistently prefers walking gaits that are either more dynamically or statically stable. These considerations may inform the synthesis of new exoskeleton gaits that maximize user comfort.

% The efficiency  
% These contributions provide insight into exoskeleton walking gaits that maximize user comfort, and 

% % that would benefit the rehabilitation community and ease future clinical trials. 

Future steps include conducting studies involving subjects with paraplegia, whose preferences likely differ from those of able-bodied subjects. As user preferences may change over time, creating a learning framework that accounts for these adaptations is also an important future research direction.

% \textcolor{red}{Additionally, given a large set of preference data, one could apply tensor decomposition techniques to discover invariant subspaces among the gait parameters. Such knowledge could accelerate learning of personalized gaits by guiding exploration.}

\dimalgo's high-dimensional learning capabilities provide insight into exoskeleton walking gaits that maximize user comfort, paving the way for generating new gaits beyond the gait library. This presents promising advancements for clinical trials and the broader rehabilitation community.

%This work towards recovering the underlying utility function of exoskeleton users provides promising avancements to the rehabilitative field.

% In addition, \dimalgo~could be adapted to optimize over user utility functions rather than gait parameters, learning personalized cost functions that generate comfortable gaits tailored to individual users.              %|
% ***********************************************

%%%%%%%%%%%%%%%%%%%%%%%%%%%%%%%%%%%%%%%%%%%%%%%%%%%%%%%%%%%%%%%%%%%%%%%%%%%%%%%%
%\section*{APPENDIX}

\section*{ACKNOWLEDGMENT}
The authors would like to acknowledge the subjects who participated in exoskeleton testing, as well as the entire Wandercraft team that designed Atalante and continues to provide technical support for this project.

\nocite{video}
\bibliographystyle{IEEEtran}
\balance
\bibliography{IEEEabrv,references}

\end{document}